\documentclass{article} 
\usepackage[preprint]{tmlr} 

\makeatletter
\if@accepted
  \newcommand{\anonmask}[2]{#2}
\else
  \newcommand{\anonmask}[2]{#1}
\fi
\if@preprint
  \def\@startauthor{\noindent \small\bf}
  \def\name{\small\bf}
  \def\addr{\footnotesize\it}
  \def\email{\hfill\footnotesize\it}
\fi
\makeatother

\usepackage{amsmath,amsfonts}
\usepackage{graphicx}
\usepackage{booktabs,tabularx,siunitx}
\usepackage{algorithm}
\usepackage{algpseudocode}
\usepackage{placeins}      
\usepackage{subcaption}    
\usepackage{arydshln}
\usepackage{hyperref}
\usepackage{url}

\newcolumntype{Y}{>{\raggedright\arraybackslash}X}
\DeclareSIUnit{\btc}{BTC}
\DeclareSIUnit{\sat}{satoshi}
\newcommand{\CI}{\ensuremath{\,\pm\,\text{95\% CI}}}

\title{MPFlow: Learning Budgeted Max-Flow Optimization on the Lightning Network with Deep Graph Reinforcement Learning}

\author{\name Harrison Rush \email harrison@amboss.tech \\
      \addr Amboss Technologies, Green Cove Springs, USA
      \AND
      \name Vincent Davis \email vincentmdavis@protonmail.com \\
      \addr Amboss Technologies, Green Cove Springs, USA
      \AND
      \name Simone Antonelli \email simone.antonelli@cispa.de \\
      \addr CISPA Helmholtz Center for Information Security
      \AND
      \name Vikash Singh \email vikash@stillmark.com \\
      \addr Stillmark, San Francisco, USA
      \AND
      \name Jesse Shrader \email j@amboss.tech \\
      \addr Amboss Technologies, Green Cove Springs, USA
      \AND
      \name Emanuele Rossi \email emanuele.rossi1909@gmail.com \\
      \addr Amboss Technologies \& Sapienza University of Rome, Barcelona, Spain
}

\def\month{MM}  
\def\year{YYYY} 
\def\openreview{\url{https://openreview.net/forum?id=XXXX}} 

\begin{document}

\maketitle

\begin{abstract}
We address liquidity placement in the Bitcoin Lightning Network (LN): given a fixed budget, which channels should a node open to maximize its routing capacity? We cast this as a budget-constrained combinatorial optimization problem on graphs, selecting $k$ edge additions that maximize $s$--$t$ max-flow, a theory-grounded measure of routing capacity, and solve it with graph reinforcement learning. Our lightweight agent combines a message-passing policy network with proximal policy optimization (PPO) and action masking, and is trained under a \emph{hub-exclusion curriculum}: the network's top hubs are removed from training subgraphs, forcing the policy to learn capacity-aware placement rather than hub attachment. In extensive experiments on real Lightning Network snapshots, our method consistently outperforms strong heuristic baselines on the max-flow objective across multiple seeds and unseen graphs. The agent has been deployed in production for peer recommendations, executing 4{,}640 channel-open decisions that cumulatively allocate 267.3 BTC (over \$16 million) across 30 managed nodes.

\end{abstract}

\begin{figure}[!hb]
  \setlength{\textfloatsep}{8pt}
  \centering
  \includegraphics[width=0.92\textwidth]{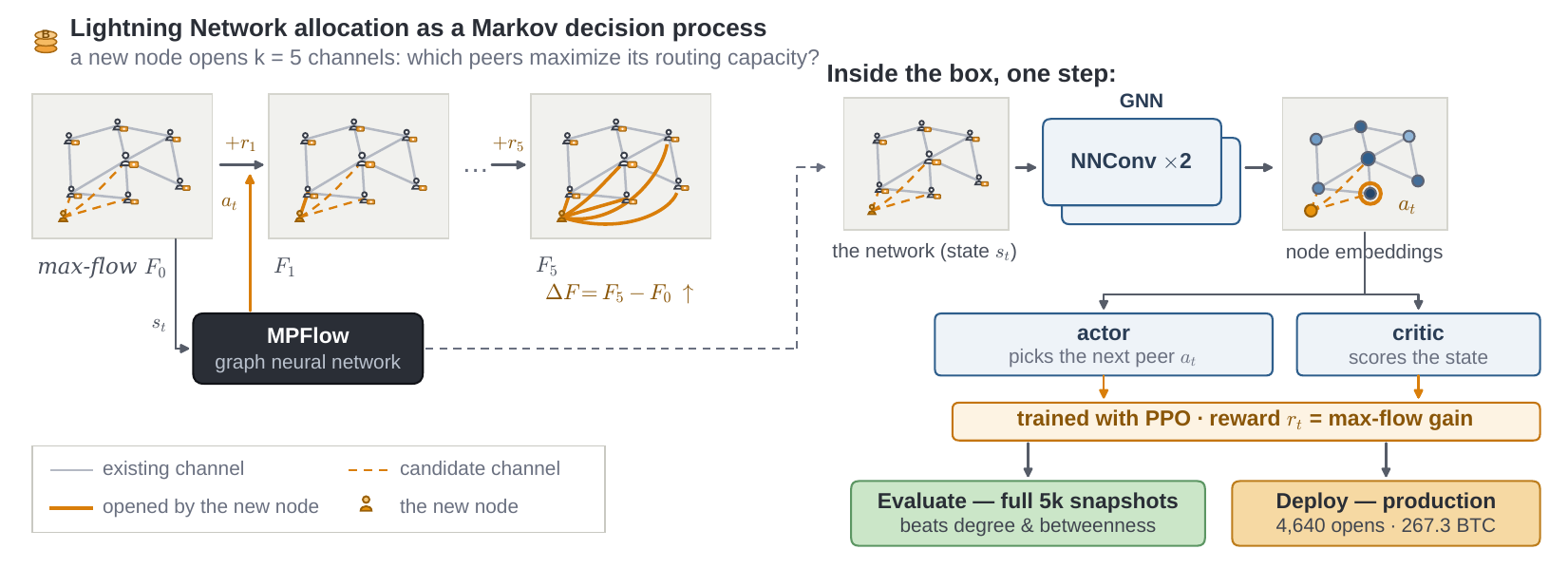}
  \vspace{-4pt}
  \caption{\textbf{Liquidity placement as learned sequential graph construction.} \emph{Left}: \textsc{MPFlow} maps each state $s_t$ to a peer $a_t$ (budget $k{=}5\times\SI{0.20}{\btc}$); reward: marginal max-flow $r_t = F_t - F_{t-1}$. \emph{Right}: two max-aggregation MPNN layers, a masked-softmax actor, and a max-pooled critic, trained with PPO.}
  \label{fig:overview}
\end{figure}

\section{Introduction}
Payment-channel networks such as the Bitcoin Lightning Network (LN) enable fast, low-cost payments by moving transactions off-chain. Their performance, however, hinges on \emph{where} liquidity is placed: poor placement creates bottlenecks that throttle routing capacity, while targeted placement can unlock throughput. For LN operators the question is simple to state and hard to solve: \emph{which peers should I connect to, and how should I allocate scarce liquidity to participate effectively?} Existing practice leans on static graph heuristics (e.g., degree or betweenness) that ignore directed balances, budgeted interventions, and interactions among multiple allocations \citep{newman2010networks,freeman1977centrality}.

The LN offers little external observability and no supervised traces; realistic traffic simulation demands strong assumptions about demand matrices, retry logic, and hidden balances that can dominate conclusions (see, e.g., LN measurement/topology studies) \citep{seres2020topology,rohrer2019discharged}. We therefore frame liquidity placement as \emph{budgeted combinatorial optimization on a graph}: select $k$ edge additions that maximize \emph{$s$--$t$ max-flow} on the observed capacity graph \citep{ford1956maxflow,ahuja1993network}, a theory-grounded objective that directly measures routing capacity and upper-bounds deliverable payment volume. In this we follow the learning-to-optimize-on-graphs line initiated by \citet{dai_learning_2018}, which trains graph-embedding RL policies to construct combinatorial solutions incrementally; in our case, the task is budget-constrained network design on a real, heavy-tailed financial network.

We instantiate a simple, deployable policy that combines a message-passing neural network (MPNN) with an on-policy optimizer (PPO) and feasibility-aware action masking. The MPNN uses a permutation-invariant \emph{max} aggregator to emphasize narrow-cut structure in heterogeneous-capacity neighborhoods \citep{gilmer_mpnn_2017}, and PPO provides stable, sample-efficient updates without second-order steps \citep{schulman_proximal_2017}. We evaluate under a paired protocol on real LN snapshots and report both \emph{seen-graph} performance and \emph{generalization to unseen snapshots}. Figure~\ref{fig:overview} summarizes the setting and the approach.

\paragraph{Contributions.}
\begin{itemize}
  \item \textbf{Optimization-aligned RL formulation:} liquidity placement as budgeted combinatorial optimization on graphs \citep{dai_learning_2018}, cast as a graph decision process with reward defined by the \emph{marginal improvement in max-flow} on observed capacity graphs \citep{ahuja1993network}.
  \item \textbf{Simple, deployable policy:} a compact MPNN--PPO agent with action masking and max aggregation \citep{gilmer_mpnn_2017,schulman_proximal_2017}.
  \item \textbf{Hub-exclusion curriculum:} we train on subgraphs with the top-50 hubs removed while evaluating on the full graph. The resulting policy allocates largely independently of top hubs (App.~\ref{app:actions}), outperforms an identically configured variant trained with hubs visible (App.~\ref{app:ablations}), and its advantage over heuristics widens as hubs are removed.
  \item \textbf{Empirical evidence on real LN topology:} consistent improvements over strong heuristic baselines \emph{on the max-flow objective} under paired evaluation, with robustness checks and multi-seed multi-snapshot results (cf. LN topology variability \citep{seres2020topology}).
  \item \textbf{Practical relevance:} the approach has been \emph{deployed in practice} to inform peer recommendations.
\end{itemize}

\section{Background}\label{sec:background}
\subsection{Bitcoin \& Lightning Network}
Bitcoin's base layer \citep{nakamoto_bitcoin_nodate} faces well-known scalability limits: low transaction throughput, slow confirmation, and high fees under congestion. The Lightning Network (LN) is a second-layer protocol that enables fast, low-cost payments through off-chain payment channels \citep{poon_bitcoin_nodate}: funds are locked in a multi-signature address, participants exchange updates privately, and only channel opening and closing touch the chain. Linking many channels yields a peer-to-peer network that routes payments between any two nodes, but each channel's fixed capacity means insufficient liquidity can block routes, and the continuously changing topology forces nodes to manage liquidity actively.

\subsection{Max-Flow}
We consider a directed graph $G=(V,E)$ with capacities $c: E \to \mathbb{R}_{\ge 0}$, source $s\in V$, and sink $t\in V$. A flow is a function $f:E\to\mathbb{R}_{\ge 0}$ satisfying capacity constraints $0\le f(u,v)\le c(u,v)$ and flow conservation $\sum_{u} f(u,v)=\sum_{w} f(v,w)$ for all $v\in V\setminus\{s,t\}$. The \emph{value} of the flow is $|f|=\sum_{w} f(s,w)=\sum_{u} f(u,t)$. The \emph{maximum flow} problem seeks $f^\star$ maximizing $|f|$. By the Max-Flow Min-Cut theorem, the maximum flow equals the minimum $s$--$t$ cut capacity. A fuller treatment, including the classical algorithms, is given in App.~\ref{app:maxflow}.

\subsection{Reinforcement Learning (MDP/PPO)}
An episodic Markov Decision Process (MDP) is $(\mathcal{S},\mathcal{A},P,r,\gamma)$ with states $s\!\in\!\mathcal{S}$, actions $a\!\in\!\mathcal{A}$, transition kernel $P$, reward $r$, and discount $\gamma$. Policy-gradient methods optimize $J(\theta)=\mathbb{E}_{\pi_\theta}\!\left[\sum_t \gamma^t r_t\right]$. Proximal Policy Optimization (PPO) maximizes a clipped surrogate objective using on-policy trajectories with an advantage estimator; see \citet{schulman_proximal_2017} for details. 

\section{Related Work}
The \emph{Lightning Network Daemon} (LND) ships an Autopilot module that opens channels automatically using heuristics such as degree and betweenness centrality \citep{noauthor_lndautopilotagentgo_nodate}; C-lightning and Eclair expose similar functionality through plugins and third-party tools \citep{folkson_plugging_2020}. Lightning Labs' production tooling (Lightning Terminal ``AutoOpen'') likewise scores new peers by betweenness centrality \citep{lnd_autoopen_docs, lnd_metrics_docs}. This reflects current industry practice for channel selection and motivates Betweenness as our primary baseline.

\citet{pickhardt_richter_optimally_2021} model path reliability by treating channel balances as random variables and computing most-likely, low-cost multi-part payments via (generalized) min-cost flow; they also argue for zero base fees to linearize the objective. Complementary analyses formalize payment success under uncertain balances and derive bounds/estimators for end-to-end success \citep{pickhardt_security_privacy_2021,pickhardt_upper_bound_upper_2021}. Closer to our data model, \citet{vincent_cbi_2024,vincent_cbi_icbc_2025} propose \emph{channel balance interpolation} (CBI), predicting local/remote splits from node/channel features; such priors can reduce variance in simulation and improve routing objectives. Our work differs by optimizing \emph{absolute max-flow uplift} with a learned MPNN--PPO policy, but shares the premise that directed balances and bottlenecks, rather than raw centrality, govern deliverability.

\citet{wang_flash_2019} proposed Flash, a dynamic routing algorithm that uses modified max-flow for large payments and routing-table lookups for small ones; \citet{sivaraman_routing_2018} proposed the Spider Network, a high-throughput routing scheme built on a congestion-control algorithm inspired by max-flow optimization; and \citet{yang_graph_2024} demonstrated a GNN-based PPO framework that adjusts node placements toward throughput objectives.

Methodologically, our approach belongs to the line of work on learning combinatorial optimization over graphs initiated by \citet{dai_learning_2018}, in which an RL policy over graph embeddings constructs a solution incrementally (there for vertex cover, max-cut, and TSP). We instantiate this paradigm for budgeted flow-network design on a real financial network, with a max-aggregation MPNN matched to the min-cut structure of the objective.

Prior work thus targets routing algorithms, balance estimation, and general network optimization; none applies deep graph reinforcement learning to budgeted liquidity allocation on the Lightning Network, the gap we address.

\section{Method}\label{sec:method}
\subsection{Problem Formulation (MDP)}
The optimization problem is: given a graph $G$ with edge capacities, a designated source, and a budget of $k$ channel openings of fixed size, choose the $k$ edge additions that maximize the resulting $s$--$t$ max-flow. Following the learned-construction paradigm for combinatorial optimization on graphs \citep{dai_learning_2018}, we build solutions incrementally and formulate the problem as a finite-horizon Markov Decision Process (MDP).
\begin{itemize}
    \item \textbf{State ($S_t$):} The state encodes the current graph topology $G_t$ including node features (centrality metrics) and edge features (fees, capacities).
    \item \textbf{Action ($A_t$):} The action space consists of selecting a node in the network to open a channel with (or top-up an existing one). We enforce a budget of $k$ actions per episode.
    \item \textbf{Reward ($R_t$):} The objective is the network's $s$--$t$ max-flow. We define the reward as the \emph{marginal} improvement in absolute max-flow:
    \[
    r_t = F_t - F_{t-1},
    \]
    where $F_t$ is the sum of max-flows calculated from the source to a set of target nodes.
\end{itemize}

\paragraph{Why max-flow as the objective.}
Our training and evaluation are anchored on $s$--$t$ Max-Flow (equivalently, Min-Cut). It is theory-grounded: by Max-Flow Min-Cut it exactly measures the structural routing capacity between the source and its targets, and upper-bounds deliverable payment volume. It is also cheap: push–relabel (or equivalent) solves in polynomial time even on large graphs, allowing us to recompute reward signals at every step. The alternative, optimizing simulated payment success directly, requires strong assumptions (arrival processes, path selection rules, fee dynamics, MPP splitting behavior) that can dominate conclusions, and is computationally prohibitive inside a training loop. Max-flow therefore serves as a theory-grounded proxy for deliverable throughput; validating the transfer to realized payment success under a credible demand model is future work (\S\ref{sec:conclusion}).

\paragraph{On the fixed budget of 5 actions.}
Real-world operators may allocate liquidity with larger or continuous budgets. In internal sweeps over \(K \in \{1, 3, 5, 10, 20\}\), \(K{=}5\) yielded the most reliable policy gradients and clearest throughput signal, so we fix \(K{=}5\) throughout and leave richer budgets to future work.

\subsection{Agent Architecture: MPFlow}
\label{sec:method_ppomn}

We adapt the standard \textsc{MPNN--PPO} framework \citep{gilmer_mpnn_2017,schulman_proximal_2017} to operate directly on the Lightning Network (LN) topology, where each edge encodes a bidirectional payment channel with known capacity and policy attributes. The agent’s objective is to allocate liquidity over visible nodes to maximize network throughput, measured as the marginal change in global max-flow. Extended architecture and training-loop diagrams appear in App.~\ref{app:figs}.

\vspace{4pt}
\paragraph{Message-passing backbone.}
Each node $i$ is initialized with a feature vector $h_i^{(0)}$ of four structural descriptors: PageRank, capacity ratio, normalized degree, and local clustering coefficient (App.~\ref{app:features}).
At each message-passing layer $l$, node $i$ aggregates messages from its neighbors $\mathcal{N}(i)$:

\begin{equation}
m_{ij}^{(l)} = \mathrm{MLP}_m^{(l)}\!\big(h_i^{(l)} \, \| \, h_j^{(l)} \, \| \, e_{ij}\big),
\qquad
e_{ij} = [\,\text{fee}^{\text{base}}_{ij}, \, \text{fee}^{\text{rate}}_{ij}, \, c_{ij}\,],
\label{eq:message}
\end{equation}

where $e_{ij}$ encodes the channel's base fee, fee rate, and capacity (log-transformed and standardized; App.~\ref{app:features}).
Instead of mean or sum aggregation, we employ a \emph{max} operator to emphasize the single most constraining neighbor, mimicking bottleneck detection along min-cut edges:

\begin{equation}
h_i^{(l+1)} = \mathrm{MLP}_h^{(l)}\!\Big(h_i^{(l)} \, \| \,
     \max_{j \in \mathcal{N}(i)} m_{ij}^{(l)} \Big).
\label{eq:agg}
\end{equation}

This \textbf{max aggregation} (Eq.~\ref{eq:agg}) replaces the usual mean-pooling used in standard MPNNs and allows the policy to focus on locally binding capacity constraints rather than diffuse averages, acting as a differentiable bottleneck detector. We use \(L{=}2\) message-passing blocks of hidden size 64 with \(\mathrm{ReLU}\) activations, followed by layer normalization of the final node embeddings; complete model settings appear in Table~\ref{tab:mpnn_config} (App.~\ref{app:hyperparams}).

\vspace{4pt}
\paragraph{Actor--critic heads.}
After $L$ message-passing layers, the shared backbone feeds two standard output heads: the actor maps node embeddings $h_i^{(L)}$ to logits and a softmax over visible nodes yields the stochastic policy $\pi_\theta(a_t|s_t)$, while the critic pools node embeddings with \emph{global max pooling} before a linear layer estimates the state value $V_\theta(s_t)$, focusing the value estimate on the most critical graph substructures rather than an average of all node states. Full head equations are given in App.~\ref{app:mpnn_description}.

\vspace{4pt}
\paragraph{Action masking and feasibility.}
Only nodes that are both \emph{visible} and \emph{operationally feasible} are permitted as actions: a binary mask assigns zero probability to infeasible choices (self-edges, disconnected nodes, or already-visited peers) and the policy is renormalized over the remainder (App.~\ref{app:mpnn_description}). This ensures compliance with LN routing rules and keeps exploration within the deployable control surface.

\vspace{4pt}
\paragraph{Why PPO.}
In preliminary experiments, A2C \citep{mnih_asynchronous_2016}, DDPG \citep{lillicrap_continuous_2019}, and a custom discrete/continuous hybrid were brittle under our reward and topology shifts, whereas PPO's clipped surrogate and entropy regularization delivered stable, sample-efficient learning with the masked discrete action space; we adopt PPO throughout and vary only the function approximator.

\section{Experimental Details}\label{sec:experimental_method}
A full overview of the hardware and experimental setup can be found in App.~\ref{app:setup}.

\subsection{Data Processing}
\label{sec:data}

The data used in this experiment consists of three network snapshots collected from a production Lightning node operated by \anonmask{an industry Lightning Network service provider (name withheld for review)}{Amboss Technologies} on May 15th (D1), July 15th (D2), and October 2nd (D3), 2025. These snapshots include a list of all open channels, their capacities $c_{\{u,v\}}$ as well as the capacities of the associated nodes $C_u$. Other node and channel features are present but not used in our case.
Each channel is first split into directed edges then aggregated for all the edges between the same pair of nodes, summing the capacities. Because per-direction balances are not publicly observable, we sample the local balances $(y_{(u,v)}, y_{(v,u)})$ uniformly at random subject to $y_{(u,v)}+y_{(v,u)}=c_{\{u,v\}}$; the full sampling procedure is given in App.~\ref{app:setup}. From the directed edges and sampled balances we construct a weighted directed graph using the iGraph Python library \citep{csardi_igraph_2006}. For all learning experiments we extract two representative sub-graphs: a 1000-node subgraph for rapid prototyping and hyper-parameter sweeps, and a 5000-node subgraph that captures $\approx$ 90\% of total network capacity while discarding the sparse long-tail of tiny nodes.
Both sub-graphs are selected by highest-degree ranking to preserve core connectivity.

\subsection{Policies Compared}
\label{sec:policies}

\paragraph{Heuristic baselines.}
To put the learned policy into context, we compare against three non-learning baselines that require progressively more graph information:

\begin{enumerate}
  \item \textbf{Random}.  A uniformly random valid node is chosen at every step.
  \item \textbf{Degree Centrality}.  Nodes are sampled proportionally to their (scaled) out-degree.
  \item \textbf{Betweenness Centrality}.  Nodes are sampled proportionally to their betweenness scores.
\end{enumerate}

All baselines observe the same action mask as the agent, maintaining fairness in feasibility constraints.

\paragraph{Learned baselines.}
Alongside \textsc{MPFlow} we train two GNN actor--critic baselines under the identical PPO protocol, differing in the encoder: \textbf{GCN}, which uses graph-convolutional layers with mean-style neighborhood aggregation and no edge features \citep{kipf2017semi}, and \textbf{GAT}, which uses attention-weighted aggregation. The learned baselines are trained \emph{without} the hub-exclusion curriculum described below. GAT was added after the primary experiments were completed; to keep the main comparison consistent with the originally run baseline set and within the available compute budget, we report it only in the cross-snapshot study (\S\ref{sec:cross_snapshot}), where pilot seen-graph runs showed the same ordering as the main table.

\paragraph{Hub-exclusion training curriculum.}
\textsc{MPFlow} is trained on subgraphs from which the 50 highest-degree nodes have been removed (\texttt{exclude\_top\_n\_degree}=50), so the agent never observes the network's top hubs during training, while \emph{all evaluation uses the full 5k subgraph, hubs included}. The intent is to deny the policy the degenerate strategy of attaching to the few largest hubs, which centrality heuristics already implement, and thereby force it onto capacity- and bottleneck-aware structure that remains valid when hubs return at evaluation time. We quantify the behavioral effect of this curriculum in App.~\ref{app:actions} and compare against an identically configured variant trained with hubs visible in App.~\ref{app:ablations}.

\paragraph{Training Protocol}
For every training episode the agent executes \(k=5\) actions, gathers 10 trajectories, and then performs one PPO update.
Training stops after 250 episodes (\(<\!6\text{ h}\) on the reference GH200).
Early stopping is triggered if the average \(\Delta\)MaxFlow over the last 10 episodes fails to improve for 20 consecutive epochs. Each learned policy in the main study (\S\ref{sec:results_main_5k}) and the ablations (\S\ref{sec:ablation}) is a \emph{single} training run with a fixed seed (evaluation variance is captured by the paired-episode CIs), whereas the cross-snapshot study (\S\ref{sec:cross_snapshot}) retrains each policy with three seeds. A diagram of the training loop is shown in Figure~\ref{fig:ppo_app} (App.~\ref{app:figs}), with PPO settings in Table~\ref{tab:ppo_config} (App.~\ref{app:hyperparams}).

\subsection{Evaluation Protocol and Metrics}
\label{sec:eval_protocol}
For each episode we fix (source, sampled balances, PRNG seed) across methods, so every policy acts on the \emph{identical} network realization, and measure the absolute gain \(\Delta F = F^{\text{after}} - F^{\text{before}}\). Each episode allows \(k{=}5\) actions (channels) at \(0.20\,\mathrm{BTC}\) per action; metrics are recorded immediately before the first and after the final action. Unless stated otherwise, uncertainty is reported as 95\% CIs computed over episodes.

\textbf{Primary metric.}
The \emph{paired uplift vs.\ Betweenness}: the per-episode relative gain of policy $\pi$ over Betweenness (the strongest heuristic and the prevailing industry baseline for peer selection), in percent, averaged across \(n{=}1000\) paired episodes.

\textbf{Secondary metrics.}
(i) Absolute improvement $\Delta F$ in \si{\btc}, averaged across paired episodes;
(ii) win rate, the fraction of paired episodes in which $\pi$ achieves higher max-flow than Betweenness;
(iii) relative improvement over the Random baseline, used in the cross-snapshot study (\S\ref{sec:cross_snapshot}) where Random provides a snapshot-independent reference.

\section{Results}\label{sec:results}
\subsection{Main Results on the 5k Subgraph}
\label{sec:results_main_5k}

On the 5k subgraph (\(n{=}1000\) paired episodes), \textsc{MPFlow} attains the highest absolute max-flow gain, \(\Delta F=0.168\pm0.003\)~BTC, and a statistically significant \emph{paired} uplift of \(+8.59\%\pm6.20\) over Betweenness, the strongest heuristic baseline (95\% CI excludes 0); see Table~\ref{tab:main_results_random} and Fig.~\ref{fig:DEL_btc_5k_app} (App.~\ref{app:figs}). It wins \(62.3\%\) of paired episodes against Betweenness, whereas every learned and heuristic policy in turn improves substantially over the Random control (\(\Delta F=0.089\pm0.003\)~BTC). A full pairwise win-rate matrix (Table~\ref{tab:winrate_matrix}, App.~\ref{app:ablations}) shows \textsc{MPFlow} beats every other policy on at least \(62\%\) of identical sampled networks, an operationally meaningful quantity: the per-decision probability that the agent makes the better allocation.

\begin{table*}[!t]
\centering
\small
\setlength{\tabcolsep}{6pt}
\caption{Main results on the 5k subgraph (\(n{=}1000\) paired episodes). \(\Delta F\) is the absolute max-flow uplift in BTC (mean \(\pm\)95\% CI). The paired difference \(\bar d\) (in millions of satoshis; \(1\,\text{BTC}=10^{8}\,\text{sat}\)) and the paired uplift~\% are measured \emph{per episode} against Betweenness, the strongest heuristic baseline; Win-\% is the fraction of episodes in which a policy beats Betweenness on the same sampled network.}
\label{tab:main_results_random}
\begin{tabular}{@{}l
  S[table-format=1.3] @{\,$\pm$\,} S[table-format=1.3]
  S[table-format=+2.2] @{\,$\pm$\,} S[table-format=1.2]
  S[table-format=+2.2] @{\,$\pm$\,} S[table-format=1.2]
  S[table-format=2.1]@{}}
\toprule
{Policy} &
\multicolumn{2}{c}{\(\Delta F\) [BTC]} &
\multicolumn{2}{c}{\(\bar d\) vs Betw.\ [\(10^{6}\) sat]} &
\multicolumn{2}{c}{Paired uplift [\%]} &
{Win-\%} \\
\midrule
Random      & 0.089 & 0.003 & -7.42\textsuperscript{$\dagger$} & 0.25 & -50.14 & 7.27 & 2.8 \\
\hdashline
Degree      & 0.154 & 0.003 & -0.88\textsuperscript{$\dagger$} & 0.14 &  -2.72 & 5.78 & 34.6 \\
Betweenness & 0.163 & 0.003 & \multicolumn{1}{c}{--} & \multicolumn{1}{c}{--} & \multicolumn{1}{c}{--} & \multicolumn{1}{c}{--} & \multicolumn{1}{c}{--} \\
\hdashline
GCN         & 0.150 & 0.003 & -1.26\textsuperscript{$\dagger$} & 0.16 & -11.38 & 3.95 & 30.6 \\
\textbf{MPFlow} & \bfseries 0.168 & \bfseries 0.003 & \bfseries +0.55\textsuperscript{$\star$} & \bfseries 0.13 & \bfseries +8.59 & \bfseries 6.20 & \bfseries 62.3 \\
\bottomrule
\end{tabular}

\vspace{2pt}
\footnotesize \emph{Notes.} \textsuperscript{$\star$}: 95\% CI of the paired uplift excludes 0 with positive sign (beats Betweenness).
\textsuperscript{$\dagger$}: 95\% CI excludes 0 with negative sign (worse than Betweenness).
Dashed lines separate the Random control and centrality heuristics from the learned policies.
\end{table*}

\subsection{Ablation Study: Graph Scale and Hub Removal}
\label{sec:ablation}
We stress-test along two axes: subgraph size $N\in\{1\mathrm{k},2\mathrm{k},3\mathrm{k},4\mathrm{k},5\mathrm{k}\}$ and targeted hub removal, each with $n{=}100$ paired episodes against Betweenness, the best heuristic in the main study. Because LN capacity is heavy-tailed, with a few high-capacity hubs mediating a disproportionate share of feasible routes, removing the top-\(k\) highest-degree nodes (\(k\in\{0,5,10,25,50\}\), from intact topology up to severe hub loss while preserving a giant component) probes whether a policy merely chases hubs or relieves structural bottlenecks. \textsc{MPFlow} matches or exceeds Betweenness across sizes (small gains at 1k–2k, a negligible dip at 3k, and a renewed margin at 4k–5k), and its advantage widens as top hubs are pruned, consistent with bottleneck relief rather than hub chasing (Table~\ref{tab:ablation_lite}; see also the action-distribution analysis in App.~\ref{app:actions}). Full means and 95\% CIs for all methods are in App.~\ref{app:ablations}; corresponding plots are in Figs.~\ref{fig:graph_size_app} and \ref{fig:hub_removal_app}.

\begin{table}[!t]
\centering
\small
\setlength{\tabcolsep}{4pt}
\caption{Ablation summary: \textsc{MPFlow} vs Betweenness (best baseline heuristic) (\(\Delta\) over Betweenness; positive favors \textsc{MPFlow}).}
\label{tab:ablation_lite}
\begin{tabular}{@{}l
  S[table-format=+1.5] S[table-format=+1.5] S[table-format=+1.5]
  S[table-format=+1.5] S[table-format=+1.5]@{}}
\toprule
& \multicolumn{5}{c}{\textbf{Graph size }$N$} \\
\cmidrule(lr){2-6}
& {1k} & {2k} & {3k} & {4k} & {5k} \\
\midrule
$\Delta F$ [\si{\btc}] & +0.02378 & +0.01781 & -0.00157 & +0.00687 & +0.00732 \\
Relative Improvement  [pp]     & +6.77    & +6.63    & -0.64    & +3.46    & +4.38    \\
\midrule
& \multicolumn{5}{c}{\textbf{Targeted hub removals }(count)} \\
\cmidrule(lr){2-6}
& {0} & {5} & {10} & {25} & {50} \\
\midrule
$\Delta F$ [\si{\btc}] & +0.00364 & +0.00479 & +0.00412 & +0.01054 & +0.00745 \\
Relative Improvement [pp]     & +2.16    & +3.19    & +3.07    & +8.89    & +7.69    \\
\bottomrule
\end{tabular}
\end{table}

\subsection{Cross-Snapshot Generalization}
\label{sec:cross_snapshot}

To evaluate temporal robustness, we train each policy on one Lightning Network snapshot and test on a later one, yielding two forward cross-time pairs: D1→D2 and D2→D3 (Table~\ref{tab:cross_snapshot_rel}). We evaluate forward pairs only: testing on a snapshot that \emph{predates} the training data would leak information about the network's earlier state into training, and does not reflect deployment, where a trained policy is always applied to future topology.

MPFlow achieves the strongest relative improvement on both pairs (D1$\to$D2: +90.6\%~$\pm$~5.3, D2$\to$D3: +99.5\%~$\pm$~6.1), demonstrating consistent generalization as the network evolves over time. Betweenness remains the strongest heuristic but trails MPFlow on both pairs; the other learned baselines are less consistent (GCN trails Betweenness on D1$\to$D2 but exceeds it on D2$\to$D3, while GAT is comparable to Betweenness on both). Overall, these results indicate that MPFlow transfers reliably to subsequent, unseen Lightning Network snapshots under topological drift.

\begin{table*}[!t]
\centering
\small
\setlength{\tabcolsep}{6pt}
\renewcommand{\arraystretch}{1.2}
\caption{Cross-snapshot generalization (\(n{=}3\) seeds; 250 paired episodes per cell). Values are mean \(\pm\)95\% CI of relative max-flow improvement (\%) over the Random baseline (ratio-of-means; bootstrap CI). Train\(\to\)Test pairs probe forward temporal drift.}
\label{tab:cross_snapshot_rel}
\begin{tabular}{@{}l
  S[table-format=+3.3] @{\,$\pm$\,} S[table-format=2.3]
  S[table-format=+3.3] @{\,$\pm$\,} S[table-format=2.3]
  @{}}
\toprule
{Policy} &
\multicolumn{2}{c}{D1 \(\to\) D2 (May \(\to\) Jul)} &
\multicolumn{2}{c}{D2 \(\to\) D3 (Jul \(\to\) Oct)} \\
\cmidrule(lr){2-3}\cmidrule(lr){4-5}
 & \multicolumn{2}{c}{Rel. Improvement [\%]} &
   \multicolumn{2}{c}{Rel. Improvement [\%]} \\
\midrule
\textbf{Random (BL)} & \multicolumn{2}{c}{--} & \multicolumn{2}{c}{--} \\
[2pt]\hdashline[0.5pt/2pt]\addlinespace[2pt]
Degree      & 73.914 & 5.218 & 85.786 & 5.722 \\
Betweenness & 85.121 & 4.997 & 88.411 & 5.734 \\
[2pt]\hdashline[0.5pt/2pt]\addlinespace[2pt]
GCN         & 76.801 & 5.257 & 92.942 & 5.525 \\
GAT         & 87.849 & 5.255 & 89.204 & 5.918 \\
\textbf{MPFlow} & \bfseries 90.558 & \bfseries 5.280 & \bfseries 99.512 & \bfseries 6.149 \\
\bottomrule
\end{tabular}
\end{table*}

\subsection{Discussion}
\paragraph{Why \textsc{MPFlow} wins on max-flow.}
A caveat first: \textsc{MPFlow} and the GCN baseline differ in several ways at once (max vs.\ mean aggregation, edge-conditioned messages, access to edge features, and the hub-exclusion curriculum of \S\ref{sec:policies}), so the mechanisms below are \emph{consistent} with our results but not isolated by them; single-factor ablations are the clearest follow-up, and the curriculum is the one factor we can already isolate (App.~\ref{app:ablations}).

Part of the margin over heuristics is objective alignment: Degree and Betweenness optimize static surrogates that ignore directed balances, budget constraints, and global flow interactions, whereas PPO optimizes an estimate of the max-flow improvement itself. But alignment alone is not sufficient: the GCN policy is trained on the identical reward yet trails Betweenness (Table~\ref{tab:main_results_random}). The expressivity we credit is the \emph{max} inductive bias. Mean aggregation averages high-capacity, low-fee neighbors together with weak or irrelevant ones, losing contrast precisely where flows are constrained by bottlenecks, whereas element-wise max aggregation preserves extreme features in each receptive field, matched to an objective whose value is governed by min-cut structure; the critic's global max pooling applies the same principle at the readout, tying the value estimate to the most constraining substructure. Whether these components jointly implement anything like min-cut reasoning \citep{velickovic_nar_2021} is speculation we do not test; we also note the GCN baseline lacks edge features, so part of the gap may reflect information access rather than aggregation. The action distributions make the two strategies concrete (App.~\ref{app:actions}): GCN places 56\% of its allocations on top-50 hubs and 47\% on just five peers (median chosen peer: degree rank 9), a learned, steeper degree heuristic, whereas \textsc{MPFlow} allocates far lower in the degree hierarchy (median rank 312; 10\% on its five most-frequent peers), consistent with relieving source-side cuts rather than attaching to hubs.

\paragraph{Robustness and the curriculum.}
Centrality heuristics concentrate their scores on hubs, so targeted hub removal degrades the very signal they rank by; \textsc{MPFlow}'s paired margin over Betweenness instead \emph{widens} as hubs are pruned (Table~\ref{tab:ablation_lite}) and persists across graph scales (Figs.~\ref{fig:hub_removal_app}, \ref{fig:graph_size_app}). This is consistent with the action distributions: only 17\% of \textsc{MPFlow}'s allocations go to top-50 hubs, versus 52\% for Betweenness (App.~\ref{app:actions}), so pruning those hubs removes far less of its preferred action set. We attribute much of this hub-independence to the training curriculum (\S\ref{sec:policies}): \textsc{MPFlow} never observes the top-50 hubs during training, and at evaluation it largely ignores them even though they are available. The curriculum also improves performance outright: in the hub-removal study, an identically configured MPNN trained \emph{with} hubs visible was evaluated on the same paired episodes, and the curriculum model achieved higher mean uplift in all four removal conditions, winning 56--62\% of paired episodes with paired gains of +1.0\% to +4.6\% (significant at the 95\% level in two of the four; App.~\ref{app:ablations}). A systematic sweep over exclusion depth, and a full-graph (no-removal) head-to-head at scale, are natural follow-ups.

\section{Conclusion}\label{sec:conclusion}
We posed liquidity placement on the Lightning Network as budgeted combinatorial optimization on a graph, selecting $k$ edge additions to maximize $s$--$t$ max-flow, and trained a compact PPO agent with an MPNN-Max backbone to construct solutions incrementally under a fixed per-action budget. On a 5k-node subgraph and 1000 paired episodes, \textsc{MPFlow} achieved the largest absolute gain and a statistically significant paired uplift over the strongest heuristic baseline, Betweenness. The policy’s advantage persists across graph scales (1k–5k) and under targeted hub removals, suggesting that it leverages capacity-aware bottleneck structure rather than raw centrality.

For operators this matters because Betweenness is the prevailing industry heuristic for peer selection; exceeding it indicates immediate, practical benefit without requiring payment traces or heavy models. Indeed, the lightweight agent (two message-passing layers, masked policy head) is already \emph{deployed in production}, powering peer recommendations on \anonmask{a commercial Lightning Network platform (name withheld for review)}{the Amboss platform}: to date, the recommender has executed 4{,}640 channel-open decisions, cumulatively allocating 267.3 BTC (over \$16 million) of on-chain capital across 30 managed nodes.

Limitations include the uniform balance sampler, a fixed $k{=}5$ per-action budget, and the fact that our objective is structural: max-flow upper-bounds, but does not by itself determine, realized success rate and economic yield. Quantifying that transfer requires a payment simulator whose assumptions (demand matrices, retry logic, hidden balances, fee-aware routing) can dominate conclusions, so we leave it to dedicated future work alongside richer reward terms (fees and costs), hybrid discrete/continuous actions, and balance priors. Nonetheless, results indicate that a simple MPNN with max aggregation, trained end-to-end with PPO, can reliably outperform topology-only heuristics on real LN topology.

\clearpage
\bibliography{references}
\bibliographystyle{tmlr}

\clearpage
\appendix

\section{Additional Figures}
\label{app:figs}

\begin{figure}[p]
  \centering
  \includegraphics[height=0.84\textheight,keepaspectratio]{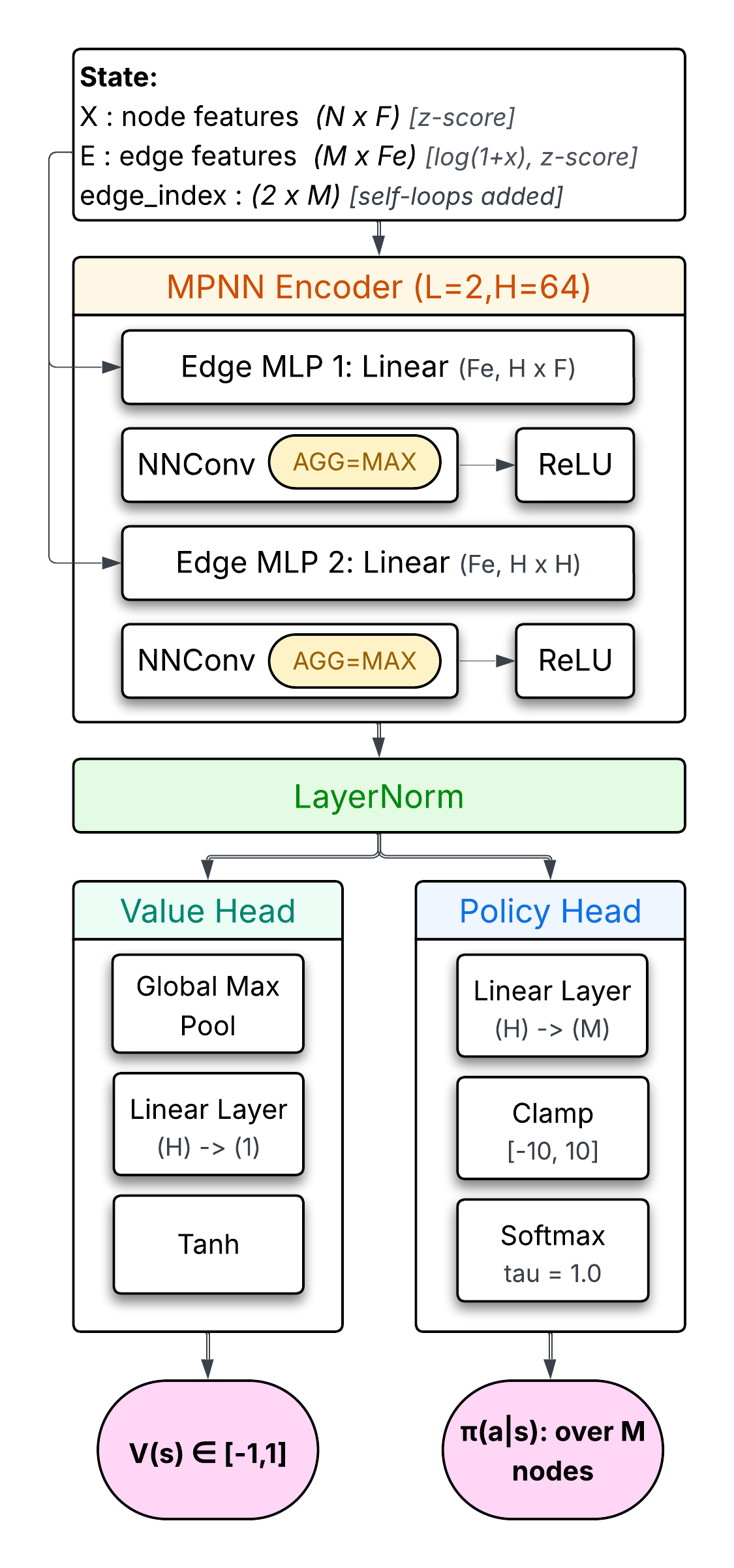}
    \caption{\textbf{MPNN--Max actor--critic used in \textsc{MPFlow}.}
    Inputs are node features $X\!\in\!\mathbb{R}^{N\times F}$, edge features $E\!\in\!\mathbb{R}^{M\times F_e}$ (log$(1{+}x)$ then z-score), and the directed edge list (self-loops added).
    The encoder has two edge-conditioned convolutions (NNConv) whose weights $W(e_{ij})$ are produced by small edge MLPs; messages are aggregated by element-wise \emph{max}, each followed by ReLU; a final LayerNorm is applied to the node embeddings.
    The actor maps node embeddings through Linear$(H\!\to\!1)$, clamps logits to $[-10,10]$, and applies a softmax over nodes to produce $\pi(a\mid s)$.
    The critic applies \emph{global max pooling} over nodes, then Linear$(H\!\to\!1)$ and $\tanh$ to output the state value $V(s)\in[-1,1]$.
    $H$ denotes the hidden width.}
  \label{fig:mpnn_app}
\end{figure}
\clearpage

\begin{figure}[p]
  \centering
  \includegraphics[height=0.6\textheight,keepaspectratio]{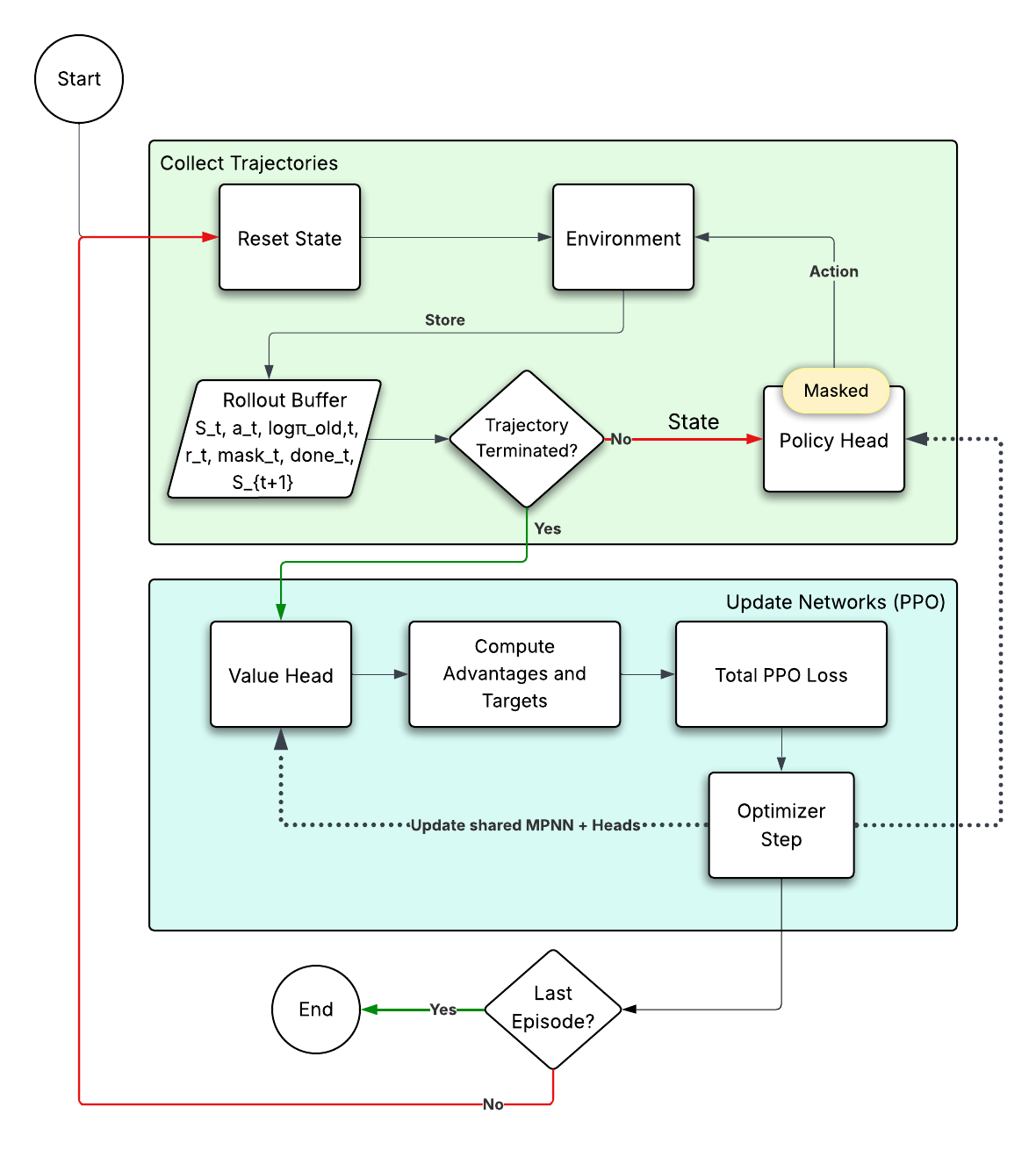}
    \caption{\textbf{PPO training flow.}
    We collect $K$ on-policy trajectories by sampling actions from a \emph{masked} policy head and append transitions $(s_t,a_t,\log\pi_{\text{old},t},r_t,\text{mask}_t,\text{done}_t,s_{t+1})$ to the rollout buffer.
    After collection, the value head (no grad) provides $V_t,V_{t+1}$ to compute TD residuals $\delta_t=r_t+\gamma(1-\text{done}_t)V_{t+1}-V_t$, advantages $A_t=\mathrm{GAE}(\delta_t;\gamma,\lambda)$, and targets $R_t=A_t+V_t$; $A_t$ is standardized.
    PPO then runs for multiple epochs and mini-batches, optimizing the clipped policy loss, value MSE, and entropy bonus with gradient clipping and Adam, updating the shared MPNN encoder and the policy/value heads.}
  \label{fig:ppo_app}
\end{figure}
\clearpage

\begin{figure}[p]
  \centering
  \includegraphics[width=\linewidth,keepaspectratio]{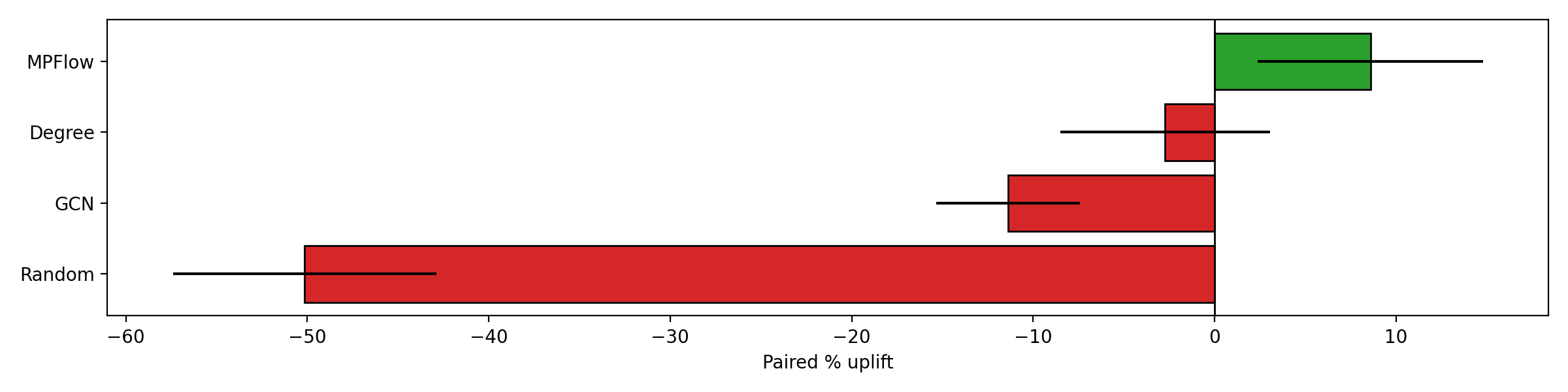}
    \caption{\textbf{Paired uplift relative to Betweenness on the 5k subgraph} (\(n{=}1000\) paired episodes; bars are mean paired \% uplift, whiskers are 95\% CIs). \textsc{MPFlow} is the only policy whose CI excludes zero on the positive side; Degree, GCN, and Random all underperform Betweenness. Values match Table~\ref{tab:main_results_random}.}
  \label{fig:DEL_btc_5k_app}
\end{figure}
\clearpage

\section{Full Ablations}
\label{app:ablations}

\begin{table}[h!]
\centering
\small
\caption{Pairwise win rates (\%) between policies on the 5k subgraph (\(n{=}1000\) paired episodes). Cell $(i,j)$ is the fraction of episodes in which row policy $i$ achieves a higher max-flow uplift than column policy $j$ on the same sampled network.}
\label{tab:winrate_matrix}
\begin{tabular}{lcccc}
\toprule
 & Random & Degree & Betw. & GCN \\
\midrule
Degree      & 96.0 & -- & 34.6 & 55.9 \\
Betweenness & 97.2 & 65.4 & -- & 69.4 \\
GCN         & 93.4 & 44.1 & 30.6 & -- \\
MPFlow     & 98.0 & 72.7 & 62.3 & 78.3 \\
\bottomrule
\end{tabular}
\end{table}

\begin{figure}[p]
  \centering

  \includegraphics[height=0.40\textheight,keepaspectratio]{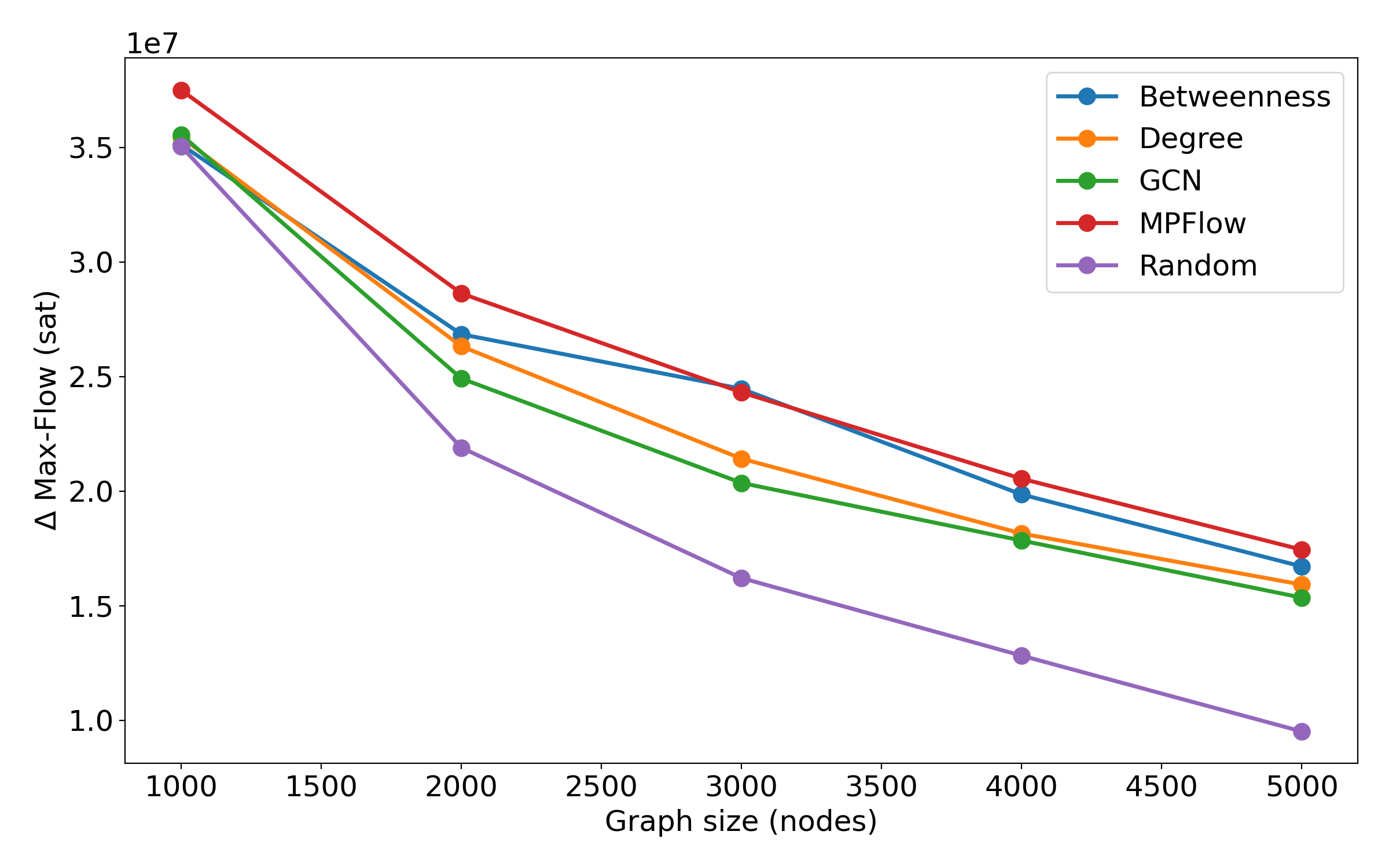}
  \caption{\textbf{Robustness under graph size.}
  Mean increase in max-flow (sat) after allocations vs.\ number of nodes included (ranked by degree).}
  \label{fig:graph_size_app}

  \vspace{0.75em}

  \begingroup
  \small
  \setlength{\tabcolsep}{4.5pt}
  \renewcommand{\arraystretch}{1.05}

  \captionof{table}{Graph size ablation: mean increase in max-flow [\si{\sat}].}
  \label{tab:ablation_graphsize_mean}
  \begin{tabular}{@{}lrrrrr@{}}
  \toprule
  Policy & 1k & 2k & 3k & 4k & 5k \\
  \midrule
  Betweenness    & 35{,}119{,}967 & 26{,}847{,}743 & 24{,}467{,}038 & 19{,}856{,}956 & 16{,}715{,}054 \\
  Degree         & 35{,}442{,}061 & 26{,}320{,}441 & 21{,}418{,}223 & 18{,}152{,}661 & 15{,}925{,}096 \\
  GCN            & 35{,}557{,}096 & 24{,}916{,}419 & 20{,}360{,}919 & 17{,}848{,}011 & 15{,}354{,}765 \\
  \textbf{MPFlow} & \textbf{37{,}498{,}380} & \textbf{28{,}628{,}957} & \textbf{24{,}310{,}135} & \textbf{20{,}543{,}831} & \textbf{17{,}447{,}526} \\
  Random         & 35{,}042{,}641 & 21{,}898{,}126 & 16{,}210{,}424 & 12{,}824{,}566 & 9{,}516{,}612 \\
  \bottomrule
  \end{tabular}

  \vspace{0.5em}

  \captionof{table}{Graph size ablation: 95\% CIs on mean [\si{\sat}].}
  \label{tab:ablation_graphsize_ci}
  \begin{tabular}{@{}lrrrrr@{}}
  \toprule
  Policy & 1k & 2k & 3k & 4k & 5k \\
  \midrule
  Betweenness    & 4{,}700{,}479 & 2{,}853{,}931 & 1{,}454{,}478 & 1{,}034{,}191 & \phantom{0}823{,}694 \\
  Degree         & 3{,}570{,}816 & 2{,}411{,}171 & 1{,}861{,}081 & 1{,}233{,}102 & \phantom{0}799{,}558 \\
  GCN            & 3{,}637{,}812 & 2{,}306{,}042 & 1{,}610{,}266 & 1{,}289{,}970 & \phantom{0}952{,}225 \\
  \textbf{MPFlow} & \textbf{3{,}584{,}004} & \textbf{2{,}638{,}046} & \textbf{1{,}709{,}296} & \textbf{1{,}207{,}376} & \textbf{\phantom{0}845{,}026} \\
  Random         & 3{,}378{,}914 & 2{,}194{,}258 & 1{,}524{,}510 & 1{,}129{,}366 & \phantom{0}845{,}773 \\
  \bottomrule
  \end{tabular}
  \endgroup
\end{figure}
\clearpage

\begin{figure}[p]
  \centering

  \includegraphics[height=0.40\textheight,keepaspectratio]{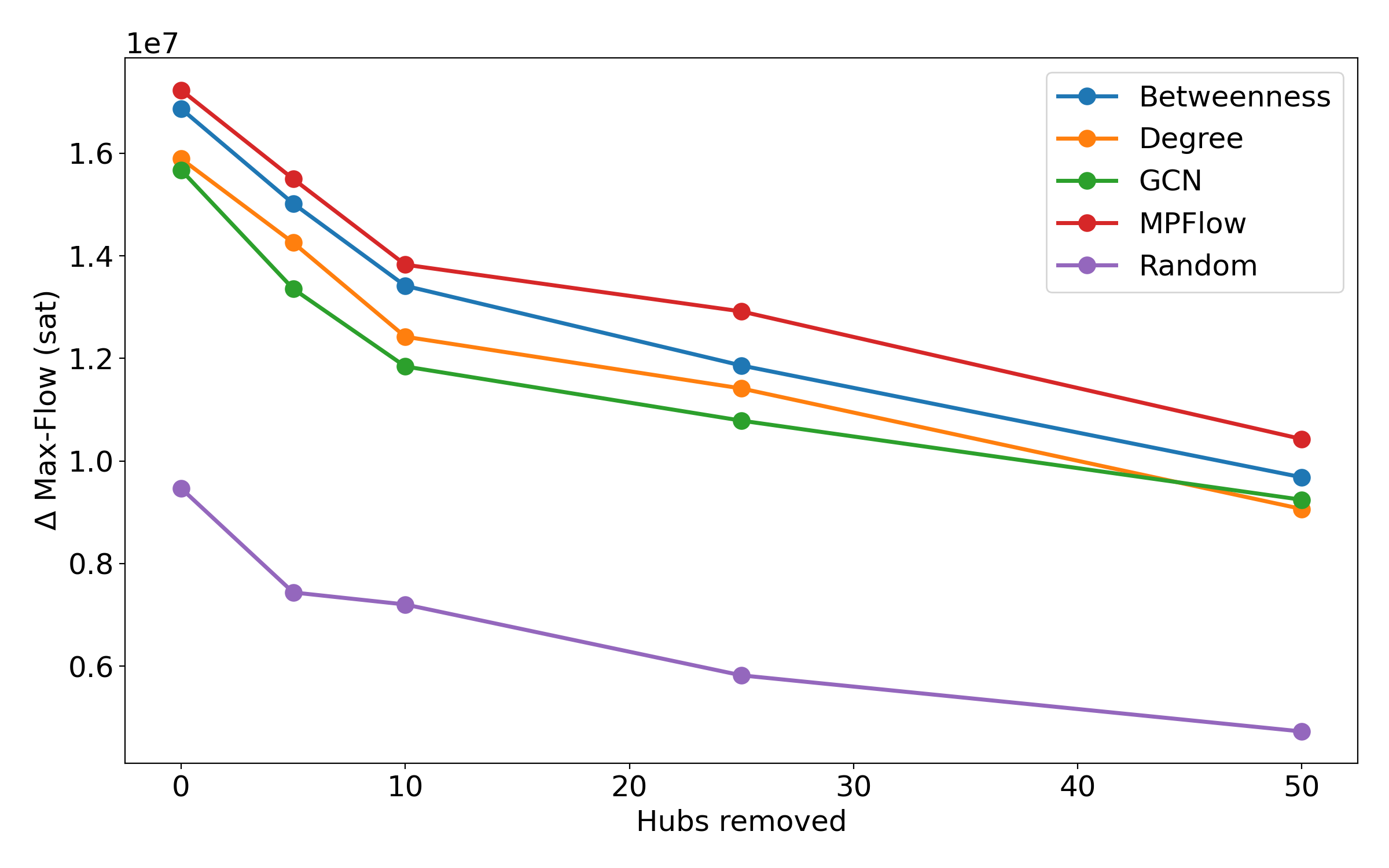}
  \caption{\textbf{Robustness under targeted hub removal (5k subgraph).}
  Mean increase in max-flow (sat) after allocations vs.\ number of highest-degree hubs removed.}
  \label{fig:hub_removal_app}

  \vspace{0.75em}

  \begingroup
  \small
  \setlength{\tabcolsep}{4.5pt}
  \renewcommand{\arraystretch}{1.05}

  \captionof{table}{Targeted hub removal: mean increase in max-flow [\si{\sat}].}
  \label{tab:ablation_hubs_mean}
  \begin{tabular}{@{}lrrrrr@{}}
  \toprule
  Policy & 0 & 5 & 10 & 25 & 50 \\
  \midrule
  Betweenness    & 16{,}863{,}171 & 15{,}016{,}857 & 13{,}413{,}550 & 11{,}860{,}529 & 9{,}682{,}913 \\
  Degree         & 15{,}889{,}267 & 14{,}253{,}300 & 12{,}421{,}288 & 11{,}413{,}222 & 9{,}064{,}480 \\
  GCN            & 15{,}668{,}308 & 13{,}354{,}329 & 11{,}841{,}622 & 10{,}785{,}778 & 9{,}244{,}112 \\
  \textbf{MPFlow} & \textbf{17{,}227{,}615} & \textbf{15{,}496{,}340} & \textbf{13{,}825{,}515} & \textbf{12{,}914{,}576} & \textbf{10{,}427{,}826} \\
  Random         & 9{,}465{,}738 & 7{,}437{,}863 & 7{,}203{,}007 & 5{,}821{,}940 & 4{,}728{,}582 \\
  \bottomrule
  \end{tabular}

  \vspace{0.5em}

  \captionof{table}{Targeted hub removal: 95\% CIs on mean [\si{\sat}].}
  \label{tab:ablation_hubs_ci}
  \begin{tabular}{@{}lrrrrr@{}}
  \toprule
  Policy & 0 & 5 & 10 & 25 & 50 \\
  \midrule
  Betweenness    & \phantom{0}825{,}513 & \phantom{0}578{,}908 & \phantom{0}637{,}288 & \phantom{0}486{,}358 & \phantom{0}508{,}064 \\
  Degree         & \phantom{0}783{,}643 & \phantom{0}666{,}313 & \phantom{0}788{,}502 & \phantom{0}455{,}548 & \phantom{0}503{,}849 \\
  GCN            & \phantom{0}849{,}652 & \phantom{0}648{,}058 & \phantom{0}709{,}782 & \phantom{0}519{,}346 & \phantom{0}514{,}983 \\
  \textbf{MPFlow} & \textbf{\phantom{0}812{,}871} & \textbf{\phantom{0}611{,}489} & \textbf{\phantom{0}812{,}458} & \textbf{\phantom{0}389{,}750} & \textbf{\phantom{0}472{,}074} \\
  Random         & \phantom{0}830{,}090 & \phantom{0}656{,}729 & \phantom{0}651{,}901 & \phantom{0}536{,}008 & \phantom{0}483{,}645 \\
  \bottomrule
  \end{tabular}
  \endgroup
\end{figure}
\paragraph{Curriculum comparison.}
The hub-removal study also evaluated \textsc{MPNN-full}, a variant with identical architecture, data, and PPO settings but trained \emph{without} the hub-exclusion curriculum (hubs visible during training). On the same paired episodes, the curriculum model (\textsc{MPFlow}) achieved higher mean uplift in all four removal conditions (Table~\ref{tab:curriculum_ablation}). Its action profile is also markedly less hub-concentrated: with 5 hubs removed, \textsc{MPNN-full} places 30\% of its allocations on top-50-degree nodes versus 9\% for \textsc{MPFlow}.

\begin{table}[h!]
\centering
\small
\setlength{\tabcolsep}{5pt}
\caption{Hub-exclusion curriculum ablation: \textsc{MPFlow} (trained with top-50 hubs excluded) vs.\ \textsc{MPNN-full} (identical model trained with hubs visible), paired per episode (\(n{=}100\) per condition, hub-removal study). \(\bar d\) is the paired difference in max-flow uplift (\(10^6\) sat, \(\pm\)95\% CI); Win-\% is the fraction of episodes the curriculum model wins.}
\label{tab:curriculum_ablation}
\begin{tabular}{@{}lrrr@{}}
\toprule
Hubs removed & \(\bar d\) [\(10^6\) sat] & Rel.\ gain [\%] & Win-\% \\
\midrule
5  & $+0.57 \pm 0.38$ & $+3.8$ & 62 \\
10 & $+0.14 \pm 0.42$ & $+1.0$ & 62 \\
25 & $+0.56 \pm 0.28$ & $+4.6$ & 61 \\
50 & $+0.13 \pm 0.25$ & $+1.3$ & 56 \\
\bottomrule
\end{tabular}
\end{table}
\clearpage
\FloatBarrier

\section{Action-Distribution Analysis}
\label{app:actions}

To test whether the policies' max-flow gains reflect hub concentration or broader capacity-aware placement, we map every action from the main study (Table~\ref{tab:main_results_random}; 5000 actions per policy over \(n{=}1000\) paired episodes) to the chosen peer's degree rank in the 5k evaluation subgraph. The subgraph topology is deterministic across episodes (per-episode balance resampling does not alter degrees), so ranks are computed once (1 = highest degree).

\begin{figure}[h!]
  \centering
  \includegraphics[width=\linewidth,keepaspectratio]{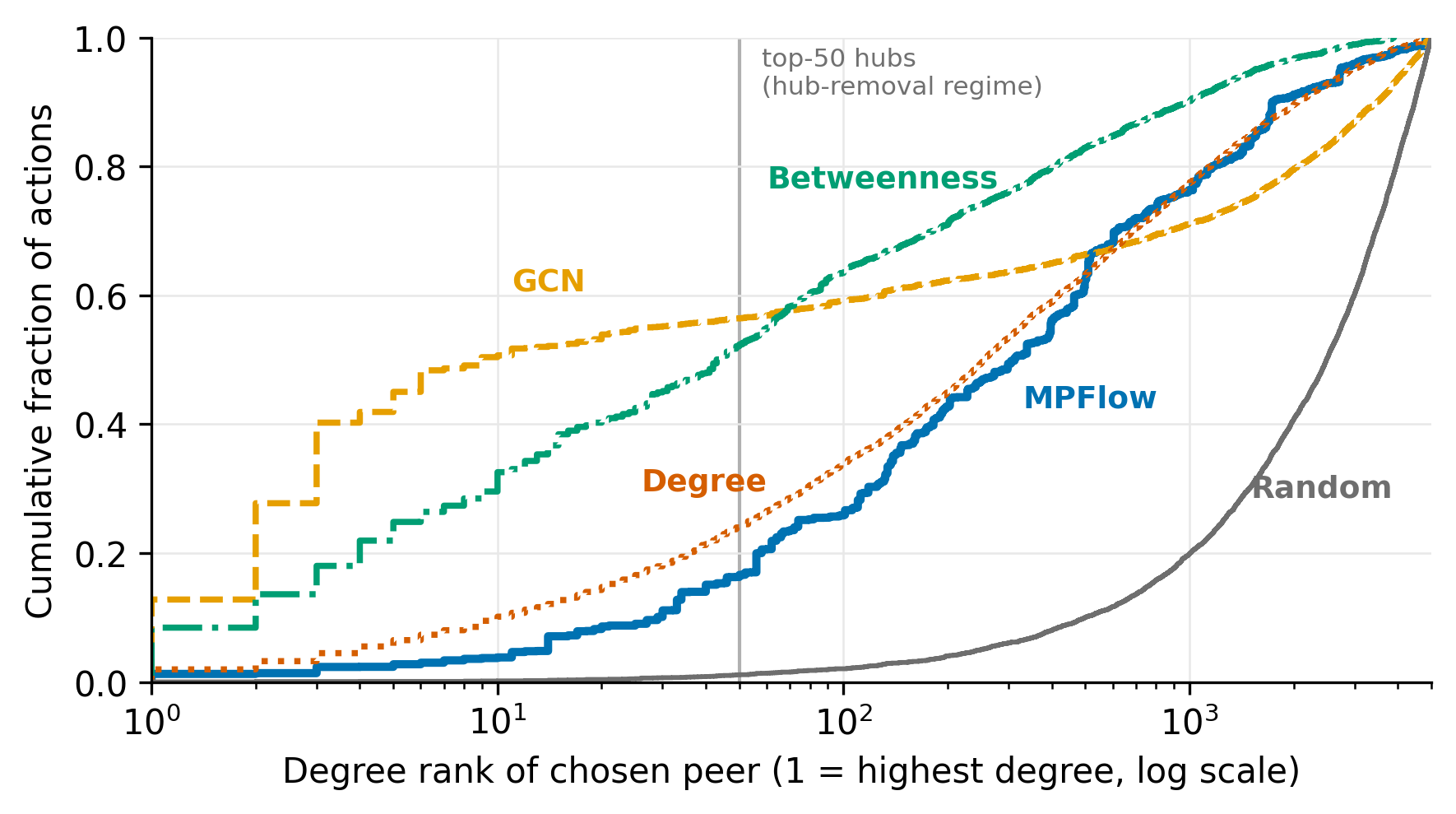}
  \caption{\textbf{Degree profile of chosen peers (main study, 5k subgraph).}
  Empirical CDF of the degree rank of the chosen peer (1 = highest degree; log scale), over all 5000 actions per policy. The vertical guide marks the top-50 nodes, the severest hub-removal regime in \S\ref{sec:ablation}. GCN and Betweenness concentrate allocations on top hubs, whereas \textsc{MPFlow}'s allocations center on mid-ranked nodes.}
  \label{fig:action_degree_app}
\end{figure}

Figure~\ref{fig:action_degree_app} and Table~\ref{tab:action_degree_app} show a clear separation. GCN and Betweenness place the majority of their allocations on top-50 hubs (56\% and 52\% of actions; median chosen peers ranked 9 and 43 by degree), and GCN concentrates 47\% of all its actions on just five peers. \textsc{MPFlow} allocates much lower in the degree hierarchy: its median pick is the 312th-ranked node, only 17\% of its actions touch top-50 hubs, and its five most-frequent peers absorb only 10\% of its actions. Random approaches the uniform diagonal, as expected.

\begin{table}[h!]
\centering
\small
\setlength{\tabcolsep}{5pt}
\caption{Action-distribution summary (5000 actions per policy). ``Top-5 conc.'' is the share of a policy's actions on its five most-frequently chosen peers.}
\label{tab:action_degree_app}
\begin{tabular}{@{}lrrrrr@{}}
\toprule
Policy & Median rank & Top-50 [\%] & Top decile [\%] & Unique peers & Top-5 conc.\ [\%] \\
\midrule
\textbf{MPFlow} & 312 & 16.6 & 62.5 & 539 & 10.1 \\
GCN              & 9   & 56.4 & 66.4 & 1624 & 46.7 \\
Betweenness      & 43  & 52.4 & 82.9 & 890 & 24.9 \\
Degree           & 256 & 24.2 & 63.2 & 1787 & 6.5 \\
Random           & 2488 & 1.1 & 10.1 & 3180 & 0.6 \\
\bottomrule
\end{tabular}
\end{table}
\clearpage

\section{Feature Definitions and Training Summary}
\label{app:features}
\paragraph{Feature definitions.}
Node features: PageRank, capacity ratio, normalized degree, local clustering; z-scored per reset. Optional edge features: base fee (msat), fee rate (ppm), channel capacity (sat) with \(\log(1{+}x)\) then z-score.

\paragraph{Training details.}
Two message-passing layers (hidden width 64), \textbf{max} aggregation. PPO with masked discrete actions; critic uses global max pooling. Additional setup (optimizer, epochs, early stopping) as in the main text; full diagrams are in Figs.~\ref{fig:mpnn_app}--\ref{fig:ppo_app}.

\section{Max-flow: full formulation}
\label{app:maxflow}
The Maximum Flow (Max-Flow) problem is a fundamental concept in network flow theory, first introduced by Lester R. Ford Jr. and Delbert R. Fulkerson in 1956. Their seminal work laid the foundation for analyzing flow networks by formulating the Max-Flow Min-Cut theorem and proposing the Ford-Fulkerson algorithm for computing the maximum flow in a network \citep{ford1956maxflow}. It involves determining the maximum amount of flow that can be sent from a source node \( s \) to a sink node \( t \) in a flow network, represented as a directed graph \( G = (V, E) \) where \( V \) is the set of vertices (nodes) and \( E \) is the set of edges (links). Each edge \( (u, v) \in E \) has a non-negative capacity \( c(u, v) \) representing the maximum flow that can pass through that edge. The problem is subject to the following constraints:
\begin{itemize}
\item\textbf{Capacity Constraints}: For every edge \( (u, v) \), the flow \( f(u, v) \) must satisfy:
\[
0 \leq f(u, v) \leq c(u, v)
\]

\item\textbf{Flow Conservation}: For every node \( u \) except the source \( s \) and sink \( t \), the sum of flows into \( u \) equals the sum of flows out of \( u \):
\[
\sum_{v \in V} f(v, u) = \sum_{v \in V} f(u, v)
\]

The objective is to maximize the total flow from the source to the sink:
\[
\text{Maximize } \sum_{v \in V} f(s, v) - \sum_{v \in V} f(v, s)
\]
\end{itemize}
Over the years, several algorithms have been developed to solve the max-flow problem more efficiently. One such algorithm is the Push-Relabel algorithm, also known as the Preflow-Push algorithm, introduced by \citet{goldberg_new_1988}. The Push-Relabel algorithm improves upon previous methods by utilizing a different approach that locally adjusts flows and maintains a preflow (a flow that allows excess at intermediate nodes) to find the maximum flow more efficiently, especially in dense networks.

\section{Experimental Setup}
\label{app:setup}
Experiments ran on an Intel 12\textsuperscript{th}-Gen Core i7 with 64\,GB RAM, an RTX\,3090 (24\,GB) and a GH200; code is in Python~3.11 using PyTorch~2.2, PyTorch-Geometric~2.5, NetworkX~3.3, and iGraph~0.10, with custom RL utilities following the \texttt{gymnasium} API. We use the LN snapshots described in Sec.~\ref{sec:data} (D1--D3); the July 2025 snapshot (D2), for reference, contains 7{,}691 nodes and 37{,}018 bidirectional channels before subgraph extraction. Each snapshot is converted into a directed, weighted graph as in Sec.~\ref{sec:data}; only total channel capacities are observed, so per-direction balances are resampled at every episode reset, and node/edge features are z-scored per reset. The \texttt{LightningNetworkEnv} extends a base graph environment: each episode permits $k{=}5$ masked liquidity actions (open/top-up); transitions deterministically update capacities; and the per-step reward is the marginal change in absolute max-flow, $r_t=F_t-F_{t-1}$, computed via push–relabel. The total max-flow reward is the sum of flows computed from our fixed source to a target set comprising 50\% of the nodes in the network, sampled randomly.

\[
S \;\sim\; \mathrm{Uniform}\bigl(\{\, U \subseteq V : |U| = \tfrac{|V|}{2} \,\}\bigr)
\]

Episodes end after the fifth action or an invalid selection; repeated max-flow solves make training largely CPU-bound.
For each undirected channel $(u,v)$ with total capacity $C_{uv}$, 
we draw a random proportion $\alpha_{uv} \sim \mathrm{Uniform}(0,1)$. 
We then assign $\alpha_{uv} C_{uv}$ to the directed edge $u \to v$ and 
the remainder $(1-\alpha_{uv}) C_{uv}$ to the reverse edge $v \to u$. 
This procedure ensures that the two directed balances $c_{uv}$ and $c_{vu}$ 
are random but always sum exactly to the channel capacity $C_{uv}$, 
providing an unbiased split of liquidity between the two directions. Each episode, this sampling is repeated.

\[
\begin{aligned}
\alpha_{uv} &\sim \mathrm{Uniform}(0,1), \\
c_{uv} &= \alpha_{uv}\,C_{uv}, \\
c_{vu} &= (1 - \alpha_{uv})\,C_{uv},
\end{aligned}
\qquad c_{uv} + c_{vu} = C_{uv}.
\]

\textbf{Uncertainty and significance.}
Unless stated otherwise, confidence intervals are \(\CI\) on means:
\(
\bar x \pm 1.96\, s/\sqrt{n}
\)
where \(x\) is either \(\Delta F\) or \(d\) and \(s\) its sample standard deviation.

\paragraph{Implementation Notes}
The MPNN backbone is implemented with \texttt{torch\_geometric.nn.MessagePassing}.  
Edge attributes are the three channel features (base fee, fee rate, capacity) defined in App.~\ref{app:features}; self-loops are added automatically.
Code and trained checkpoints are available in the accompanying repository.

For each reset of the environment we build two matrices: a node–feature matrix
\(X\in\mathbb{R}^{N\times 4}\) and (optionally) an edge–feature matrix
\(E\in\mathbb{R}^{M\times 3}\).  
All features are re-computed after any topological change and standardised
(z-score per dimension) before being fed to the MPNN.

\paragraph{Environment}
We implement a lightweight \texttt{BaseNetworkEnv} (graph ops via iGraph) and two specializations: \texttt{RandomNetworkEnv} (Barabási–Albert graphs) for smoke tests and \texttt{LightningNetworkEnv} for real LN snapshots (loading, masked open/top-up actions, deterministic capacity updates). The API mirrors \texttt{gymnasium} with custom \texttt{reset}/\texttt{step}; rewards use push–relabel as defined earlier, enabling drop-in swapping of graph sources while keeping identical RL code.

\subsection{Node and Edge Features}
\label{app:node_features}
\begin{enumerate}
  \item \textbf{PageRank Centrality} \(\;C_{PR}(u)\).  
        Captures global influence and replaces the earlier adjacency indicator:
        \[
        X_{u,1}=C_{PR}(u).
        \]

  \item \textbf{Capacity Ratio} \(\;C_{CR}(u)\).  
        Fraction of total network liquidity held by \(u\):
        \[
        C_{CR}(u)=\frac{C_u}{\sum_{v\in V} C_v}, \qquad
        X_{u,2}=C_{CR}(u).
        \]

  \item \textbf{Normalised Degree} \(\;C_{D}(u)\).  
        Relative connectivity of \(u\):
        \[
        C_{D}(u)=\frac{\deg(u)}{N-1},\qquad
        X_{u,3}=C_{D}(u).
        \]

  \item \textbf{Local Clustering Coefficient} \(\;C_{C}(u)\).  
        Density of the node’s neighbourhood:
        \[
        X_{u,4}=C_{C}(u).
        \]
\end{enumerate}

When \texttt{use\_edge\_features} is enabled the MPNN receives three attributes
per directed edge \((u,v)\):
\begin{enumerate}
  \item \textbf{Base Fee} \(\;f^{\text{base}}_{uv}\).  
        Fixed cost charged on any payment through \((u,v)\):
        \[
        X_{uv,1} = f^{\text{base}}_{uv}\,[\mathrm{msat}].
        \]

  \item \textbf{Fee Rate} \(\;f^{\text{rate}}_{uv}\).  
        Proportional fee charged per unit of forwarded liquidity:
        \[
        X_{uv,2} = f^{\text{rate}}_{uv}\,[\mathrm{ppm}].
        \]

  \item \textbf{Channel Capacity} \(\;c_{uv}\).  
        Total liquidity available on channel \((u,v)\):
        \[
        X_{uv,3} = c_{uv}\,[\mathrm{sat}].
        \]
\end{enumerate}
To mitigate heavy-tailed distributions we first apply a
\(\log(1+x)\) transform:
\[
\tilde{e}_{uv,i}=\log\!\bigl(1+e_{uv,i}\bigr)\quad(i=1,2,3),
\]
followed by per-column normalisation
\(E\leftarrow(\tilde{E}-\mu_E)\oslash(\sigma_E+\varepsilon)\).

Edges with missing fee information default to zeros, ensuring that the feature
tensor always has shape \(M\times3\). 

\subsection{MPNN-Max Architecture}
\label{app:mpnn_description}
The actor and critic share the same message-passing backbone but have separate output heads.

\begin{enumerate}
  \item \textbf{Input}.  
        Node feature matrix \(X\in\mathbb{R}^{N\times F}\) and edge feature matrix \(E\in\mathbb{R}^{M\times F_e}\).

  \item \textbf{Message Passing Block} (repeated \(L=2\) times).  
        For each edge \(e=(u,v)\) the message function is
        \[
        m_{uv}^{(l)} = \sigma\!\bigl(W_m^{(l)}[h_u^{(l)}\,\|\,h_v^{(l)}\,\|\,e_{uv}]\bigr),
        \]
        followed by node update
        \[
        h_u^{(l+1)} = \sigma\!\Bigl(W_h^{(l)}\bigl[h_u^{(l)}\,\|\,\mathrm{AGG}\{m_{vu}^{(l)}\}_{v\in\mathcal{N}(u)}\bigr]\Bigr),
        \]
        where AGG is element-wise \emph{maximum}. Max aggregation proved empirically more discriminative for heterogeneous channel capacities than mean. This design follows the Message Passing Neural Network (MPNN) framework \citep{gilmer_mpnn_2017}, with edge-conditioned convolutions in the spirit of ECC/NNConv \citep{simonovsky_ecc_2017}.

  \item \textbf{Policy Head (Actor)}.  
        A linear projection maps \(h^{(L)}\) to logits \(o\in\mathbb{R}^N\).  
        The masked soft-max produces the stochastic policy:
        \[
        \pi_\theta(a_i\mid s)=\frac{\exp(o_i)}{\sum_{j\in\mathcal{A}(s)}\exp(o_j)}\,, \qquad i\in\mathcal{A}(s),
        \]
        where \(\mathcal{A}(s)\) is the valid action set given by the environment mask.

  \item \textbf{Value Head (Critic)}.
        Global \textbf{max} pooling aggregates node embeddings:
        \(
        h_G=\max_{u\in V} h_u^{(L)}.
        \)
        A single fully-connected layer with \(\tanh\) activation outputs the state-value estimate \(V_\phi(s)\in[-1,1]\).
\end{enumerate}

\subsection{Hyperparameter Settings}
\label{app:hyperparams}
Table~\ref{tab:mpnn_config} summarizes the MPNN actor--critic configuration used for \textsc{MPFlow}; Table~\ref{tab:ppo_config} lists the PPO settings shared by all learned policies.

\begin{table}[h!]
\centering
\caption{MPNN actor–critic configuration.}
\label{tab:mpnn_config}
\begin{tabular}{@{}ll@{}}
\toprule
Parameter & Value \\
\midrule
Num node features & 4 \\
Num edge features & 3 \\
Hidden dimension & 64 \\
Num message-passing layers & 2 \\
\bottomrule
\end{tabular}
\end{table}

\begin{table}[h!]
\centering
\caption{PPO hyperparameters.}
\label{tab:ppo_config}
\begin{tabular}{@{}ll@{}}
\toprule
Parameter & Value \\
\midrule
Clip ratio \((\epsilon)\) & 0.2 \\
Value loss coefficient \((c_1)\) & 0.5 \\
Entropy coefficient \((c_2)\) & 0.01 \\
Discount \((\gamma)\) & 0.99 \\
GAE \((\lambda)\) & 0.95 \\
Learning rate \((\alpha)\) & 0.002 \\
Optimizer & Adam \\
PPO epochs & 5 \\
Minibatch size & 64 \\
Trajectories per update & 10 \\
Max grad norm & 1.0 \\
\bottomrule
\end{tabular}
\end{table}

\end{document}